\documentclass[letterpaper,10pt,conference]{ieeeconf}
\IEEEoverridecommandlockouts

\usepackage{amssymb}
\usepackage{booktabs}
\usepackage{color}
\usepackage[backend=biber,
            url=false,
            isbn=false,
            doi=false,
            backref=false,
            style=ieee,
            natbib=true,
            mincitenames=1,
            maxcitenames=1,
            citestyle=numeric-comp,
            sorting=none,
            block=none]{biblatex}
\renewcommand{\bibfont}{\small}
\usepackage{cuted}

\usepackage{graphicx}
\let\labelindent\relax
\usepackage{enumitem}
\usepackage{amsmath}
\usepackage{wrapfig}
\usepackage{multirow}
\usepackage{caption}
\usepackage{subcaption}
\usepackage{tabularx,array}
\usepackage{listings}
\usepackage{xcolor}
\usepackage{tcolorbox}
\usepackage{xspace}
\usepackage{makecell}

\usepackage[colorlinks=true,
            linkcolor=blue,
            citecolor=blue,
            urlcolor=blue]{hyperref}
\usepackage[nameinlink, capitalize]{cleveref} 

\usepackage{siunitx}
\usepackage{caption}
\newcommand{\ACRO}{{DexDirect}\xspace}

\definecolor{lightgray}{gray}{0.95}

\lstdefinelanguage{yaml}{
  morekeywords={true,false,null,yes,no},
  sensitive=false,
  morecomment=[l]{\#},
  morestring=[b]",
  morestring=[b]'
}

\begin{document}
    
\title{\ACRO: Direct Kinesthetic Arm Guidance for Efficient Dexterous Demonstration Collection}

\author{Beom Jun Kim$^{1}$, Shiu-Jen Wang$^{1}$, Jonathan Liu$^{2}$, Alvin Zhu$^{2, 3}$, Quanyou Wang$^{1}$, \\ Hanzhang Fang$^{2, 3}$, Feng Xu$^{1}$, Mingzhang Zhu$^{1}$, Yuchen Cui$^{3\dagger}$, and Dennis W. Hong$^{1\dagger}$ \\ 
\thanks{$\dagger$ denotes equal advising.}
\thanks{$^{1}$Department of Mechanical and Aerospace Engineering, $^{2}$Department of Electrical and Computer Engineering, $^{3}$Department of Computer Science, $^{4}$Department of Electrical and Computer Engineering,  UCLA, Los Angeles, CA, USA.}
}

\maketitle

\begin{abstract}
Scalable collection of dexterous manipulation demonstrations remains a major bottleneck for robot learning. High-fidelity interfaces often require costly hardware and extensive setup, while low-setup, low-cost alternatives tend to provide less precise control and impose greater cognitive workload on operators. We present \ACRO, a direct kinesthetic arm guidance for efficient dexterous demonstration collection. The operator drags a 6-DoF gravity-compensated robot arm directly by a handle, while a single webcam retargets operator's other hand onto a 16 joints 13-DoF dexterous robot hand. User studies suggest \ACRO collects 17.2$\times$ and 3.2$\times$ more successful demonstrations compared to purely vision (AnyTeleop) and pose-tracking (TeleDex) baselines. An adapted NASA-TLX shows \ACRO greatly reduces mental demand, effort, and frustration, despite raising physical demand. A diffusion policy trained on \ACRO demonstrations reaches a 90\% success rate on a cube pick-and-place task. These results suggest that direct kinesthetic arm guidance combined with vision-based hand retargeting provides an efficient low-setup and scalable interface for collecting dexterous manipulation demonstrations.
\end{abstract}


\section{Introduction}
Learning robot manipulation policies from demonstrations has become a
central paradigm for acquiring complex, contact-rich behaviors.
The performance of these policies depends critically on the scale and
quality of the collected demonstrations: an effective data-collection
interface should produce accurate, robot-executable trajectories while
remaining fast, inexpensive, and easy to deploy. In practice, however, it is difficult to achieve all of these objectives simultaneously. Interfaces that provide precise
control and rich interaction feedback typically require specialized
hardware and extensive setup, whereas lightweight alternatives often
sacrifice control fidelity or increase the burden on the operator.

For parallel-jaw grippers, this trade-off has been mitigated by mature
and scalable interfaces. Lightweight handheld devices allow
demonstrations to be collected in diverse environments~\cite{umi},
while isomorphic leader arms provide accurate and intuitive joint-space
control~\cite{gello,aloha}. Extending these solutions to multi-fingered
hands is considerably more difficult. Although anthropomorphic robot
hands resemble human hands, they differ in link geometry, joint axes,
ranges of motion, compliance, and actuation. Moreover, the operator must simultaneously command the global 6-DoF
motion of the arm and the high-dimensional articulation of the fingers,
while regulating contacts that are often perceived only through visual
feedback. These factors make dexterous demonstration collection slow,
operator-dependent, and difficult to scale~\cite{exostart,dexpilot}.

Existing interfaces address this bottleneck by trading setup cost for demonstration fidelity. At one end are hardware-based interfaces, such as exoskeletons, that instrument the operator's hand and retarget measured joint angles to the robot~\cite{dexumi}. Some additionally retain contact forces through using linkage mechanisms~\cite{ace,dexop,dexexo}. Although these systems differ mechanically, they all rely on dedicated linkage hardware: many require operator-specific fitting or calibration, while even operator-agnostic designs such as DexEXO~\cite{dexexo} must be redesigned or refabricated for a new robot.

At the other end are vision-based and pose-tracking approaches that eliminate dedicated hardware altogether~\cite{anyteleop,dexpilot,telekinesis}. A commodity webcam or depth camera is sufficient to begin collecting demonstrations, while wearable pose trackers, such as a smartphone or headset, can improve tracking quality without introducing custom mechanical interfaces~\cite{teledex,omnih2o}. We use this class of systems as the reference point for low-setup cost.

\begin{figure}[t!]
 \centering
 \includegraphics[width=\linewidth]{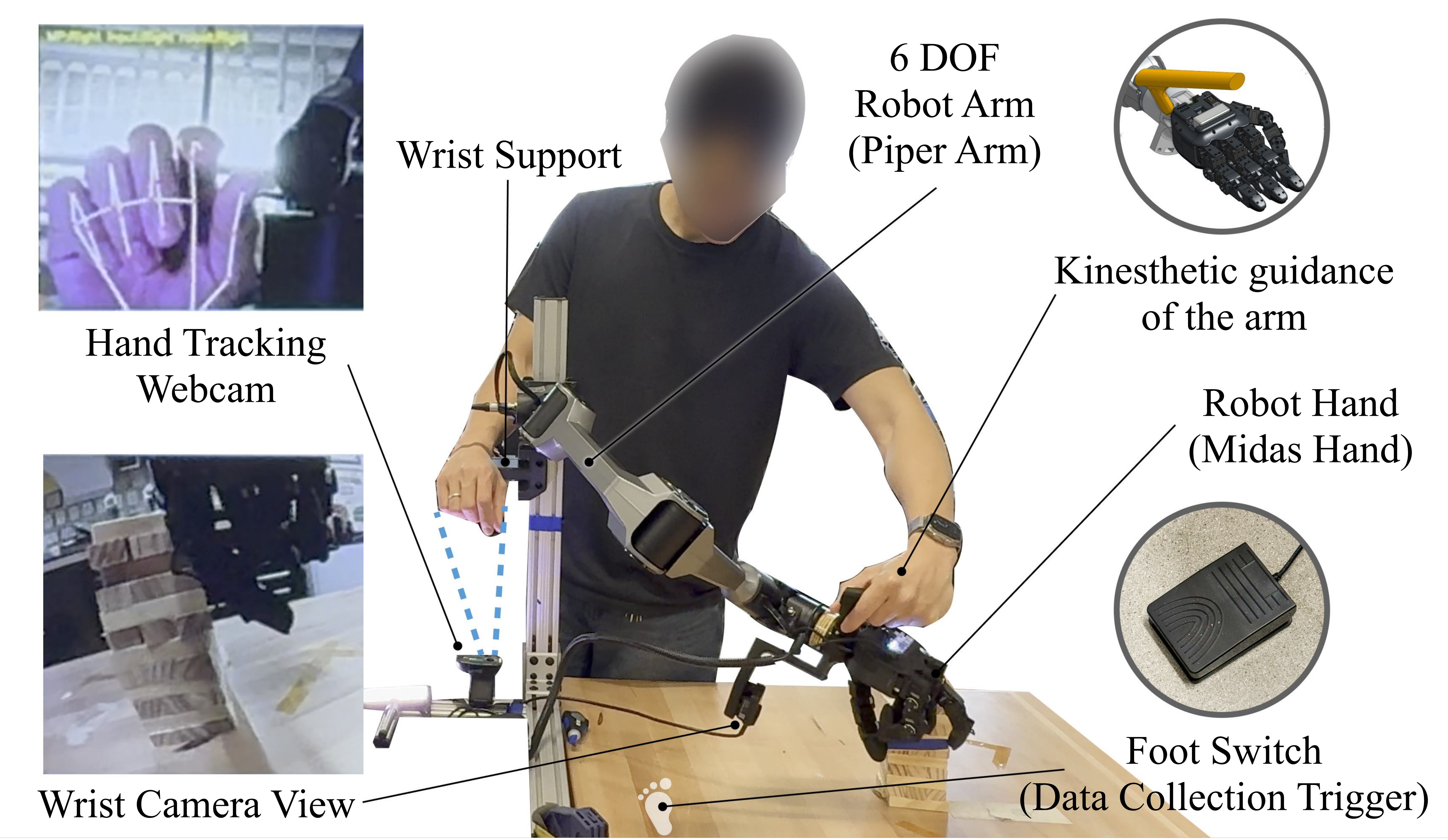}
 \caption{\ACRO combines direct kinesthetic guidance of the robot arm with vision-based retargeting of the operator's free hand. One hand guides the robot through a wrist-mounted handle, while the other commands the dexterous hand.}
 \label{fig:main}
\end{figure}

Low-setup systems, however, generally rely on a tracking-based control loop in which the operator moves a human hand or handheld device and the robot follows the estimated motion. Consequently, demonstration performance is affected by factors such as tracking latency, pose estimation error, workspace mismatch, and the cognitive burden of maintaining a coordinate mapping between the operator and the robot. 

A natural way to avoid this proxy loop is to guide the robot directly.
Kinesthetic teaching has been shown to provide an intuitive interface
for non-expert users while recording configurations physically attained
by the robot~\cite{wrede2013kinesthetic,akgun2012kinesthetic,
li2025demonstration}. Applying the same approach to dexterous fingers,
however, is non-trivial because manually guiding many small joints is
mechanically cumbersome.

This asymmetry motivates our hybrid design. We present \ACRO, a low-setup interface for efficient dexterous demonstration collection that combines direct kinesthetic arm guidance with vision-based finger retargeting. The operator guides a gravity-compensated 6-DoF robot arm through a handle rigidly attached to the robot wrist, allowing the arm trajectory to be recorded directly from the robot joint encoders rather than inferred through external tracking. At the same time, a single webcam tracks the operator's other hand and retargets it to a 16 joints 13-DoF dexterous robot hand using the same class of commodity vision pipeline adopted by prior vision-based systems. As a result, \ACRO requires no leader arm, no operator-to-robot calibration for arm teleoperation, and no arm-side retargeting, while preserving the simple deployment of webcam-based systems.

In summary, our contributions are:

\begin{itemize}
  \item \textbf{A low-setup, leader-less interface for dexterous demonstration collection.}
        \ACRO combines direct kinesthetic arm guidance with vision-based hand retargeting, eliminating separate leader hardware, arm-side tracking noise, occulusion and latency.

  \item \textbf{Higher demonstration throughput.}
         Across ten participants and five tasks, \ACRO collected more successful demonstrations, compared to TeleDex and AnyTeleop.

  \item \textbf{Lower operator workload.}
      In our user study, DexDirect reduced perceived mental demand, effort,
      and frustration, resulting in significantly lower overall workload than
      both baselines, as measured by NASA-TLX~\cite{hart1988development}.
\end{itemize}

\section{Related Work}


\subsection{Direct Physical Demonstration}
One family of interfaces collects demonstrations by physically guiding a body that shares the robot's kinematics, so the trajectory is read directly from the joint encoders without pose estimation or retargeting. Kinematically isomorphic leaders use small-scale or duplicate robot arms as leader controllers, mapping joint space directly and thereby avoiding inverse kinematics failures near singularities while implicitly teaching operators the follower's physical limits~\cite{gello,aloha,lerobot,mobilealoha}, and other works split the master device into large-range workspace navigation and fine-grained end-effector manipulation~\cite{globallocal}. A key drawback, however, is that a second leader arm nearly doubles hardware requirements, resulting in high setup cost~\cite{openteach}.

Other direct demonstration methods remove this duplicate setup by having the operator backdrive the deployment robot itself~\cite{lfd_survey}. This also carries a cost: because the operator's hands occlude the camera's view of the contact region, each trajectory must be re-executed offline~\cite{dexforce, forcematch}, or the observation must be postprocessed through inpainting or point cloud prediction~\cite{kinedex, kinesoft}.

\subsection{Exoskeleton Interfaces}
Exoskeleton interfaces instrument the operator's own limb and retarget the measured joint angles to the robot~\cite{ace,dexop,dexexo,airexo2,ume,nuexo}, with several recovering contact force directly through the linkage. Related worn hand interfaces retarget the operator's finger motion while using visual inpainting~\cite{dexumi} or a wrist camera~\cite{dexexo} to obtain policy observations for the robot hand. These interfaces exhibit a common trade-off, with mechanical linkages, servo motors, or extra instrumentation that add further hardware, calibration, and per-operator fitting.

\subsection{Vision and Pose-Based Teleoperation}
Vision and pose-based teleoperation is a cheaper alternative to expensive master-device and exoskeleton systems~\cite{teledex,anyteleop,opentelevision,bunnyvisionpro, telekinesis,dexpilot,omnih2o}. AnyTeleop~\cite{anyteleop} translates human movement to a robot using a standard webcam, pairing software-driven retargeting with a real-time motion generation module to map human keypoints to multi-fingered hands on the fly. TeleDex~\cite{teledex} further expands portability, turning a commodity smartphone into a standalone interface through a 3D-printable wrist mount for hand articulation, allowing for demonstration collection with minimal setup.

However, relying on monocular webcam vision leaves systems vulnerable to self-occlusion while inherently suffering from processing latency and tracking jitter~\cite{dexpilot}. While tracking a handheld device's own inertial and visual odometry circumvents visual occlusion~\cite{teledex}, operating a phone in free space offers zero environmental resistance. This complete absence of haptic feedback leaves force regulation unconstrained, making continuous contact-rich manipulation, dynamic target tracking, and precise spatial coordination cognitively taxing and highly susceptible to operator error~\cite{glovity}.

\subsection{Handheld Capture}
To enable real-time, single-pass collection, an alternative line of work decouples capture from the robot entirely using lower-cost wearable~\cite{dexcap,dexmouse,glovity} or handheld interfaces~\cite{umi,forcemimic}. Several of these interfaces add synthetic haptic or wrench feedback so that operators can regulate contact force during collection~\cite{dexmouse,glovity}. Although these devices can capture data in a single pass, their synthetic force feedback is governed by secondary control loops with inherent bandwidth limits, and recorded forces reflect proxy mechanics rather than the deployment manipulator's true joint inertia, motor compliance, and full wrench constraints~\cite{forcemimic,glovity}.

\begin{figure*}[t]
    \centering
    \includegraphics[width=\linewidth]{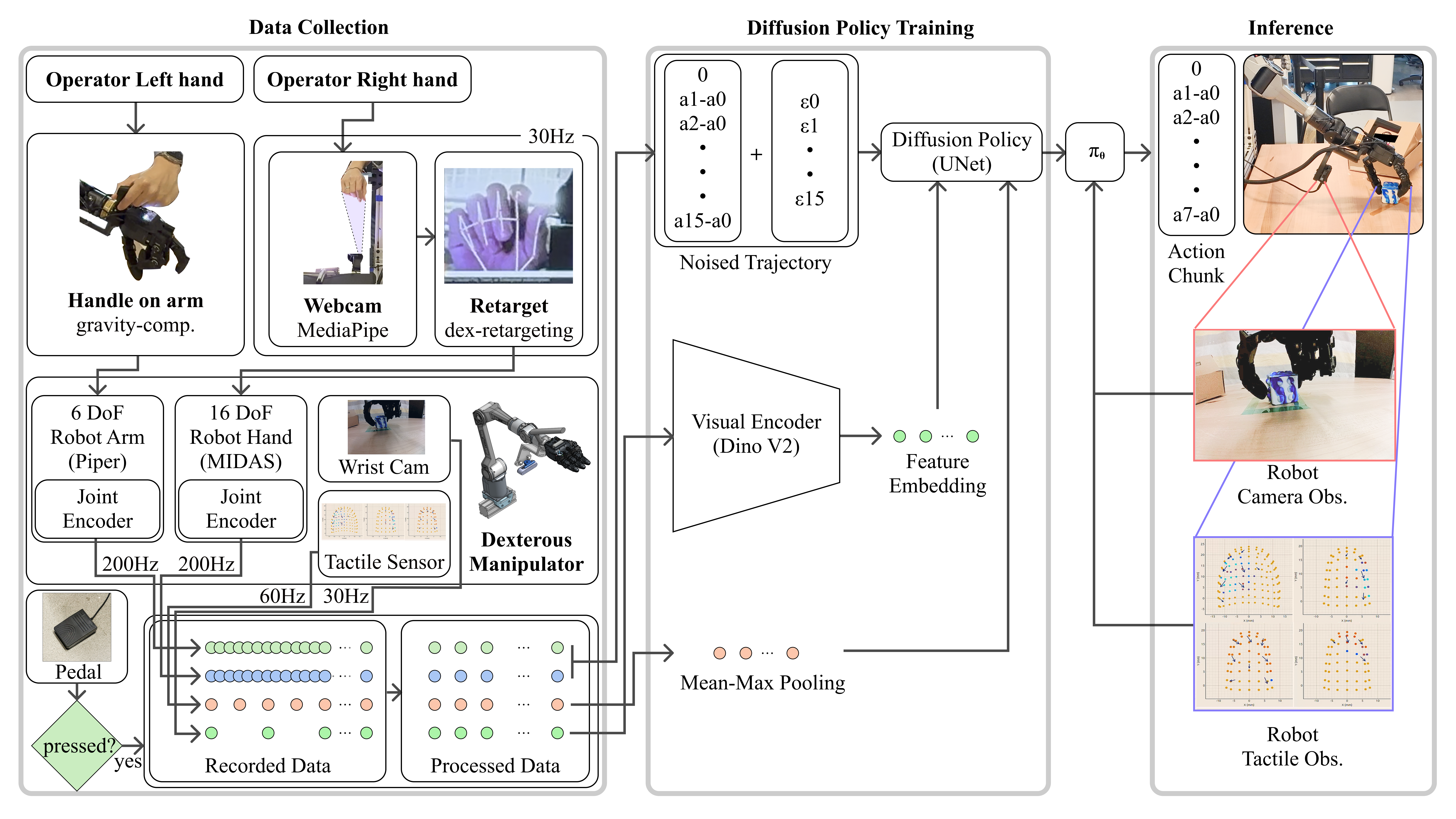}
    \captionsetup{belowskip=-10pt}
    \caption{\textbf{System overview of DexDirect}
The operator directly guides the gravity-compensated robot arm while controlling the dexterous hand through vision-based hand-pose retargeting. The collected joint, visual, and tactile observations are synchronized and used to train a multimodal diffusion policy. At inference, the policy predicts robot action chunks conditioned on the current camera and tactile observations.}
    \label{fig:system}
\end{figure*}

\section{System Design}
\label{sec:system}

\subsection{System Overview and Design Rationale}
\label{sec:system:overview}

\ACRO separates arm motion and finger articulation into two
complementary control pathways. As shown in
Fig.~\ref{fig:system}, the operator's left hand directly guides the
physical robot arm through a rigidly attached handle, while the
right hand controls the robot fingers through webcam-based hand
tracking. A multimodal recording pipeline logs the resulting robot
states, tactile measurements, and visual observations.

This separation follows from the different requirements of arm and
finger control. Arm motion requires global 6-DoF positioning over a
large workspace and involves physical interaction with the
environment. Finger articulation is high-dimensional, but can be
represented primarily by the relative configuration of the fingers
in a hand-local frame, because the robot arm independently determines
the global pose of the hand. \ACRO therefore uses direct kinesthetic
guidance for the arm and vision-based retargeting for finger
articulation. Because the arm determines the robot hand's global pose, the vision
pathway uses only wrist-relative finger geometry, enabling a low-cost,
non-wearable interface without arm-side pose tracking.

\noindent\textbf{Manipulator.}
The platform consists of an Agilex PIPER 6-DoF manipulator and a dexterous MIDAS hand~\cite{midas} mounted at the flange. The MIDAS hand has
16 joints driven by 13 actuators, is backdrivable,
and contains 283 distributed tactile taxels. The total payload mounted at the
end-effector flange is \SI{0.75}{\kilo\gram}.

\noindent\textbf{Operator handle.}
A simple 3D-printed PLA handle weighing \SI{0.05}{\kilo\gram} is mounted on
the dorsal side of the hand assembly, with its grip point offset
\SI{65}{\milli\meter} from the flange along the palm normal. This offset
provides clearance for the operator's hand while limiting the moment arm of
the applied guidance force.

\noindent\textbf{Sensing and operator station.}
A wrist-mounted RGB camera (Luxonis OAK-1 Wide FOV) records the visual
observations used for policy learning. A separate webcam (Logitech C270),
mounted \SI{45}{\centi\meter} below the operator's right hand and oriented
toward the palm, drives finger tracking. The operator stands beside the workspace
with the left hand on the handle and the right forearm resting on a
fixed 3D printed support. A nearby monitor displays the wrist-camera stream.

\begin{figure*}[t!]
 \centering
 \includegraphics[width=\linewidth]{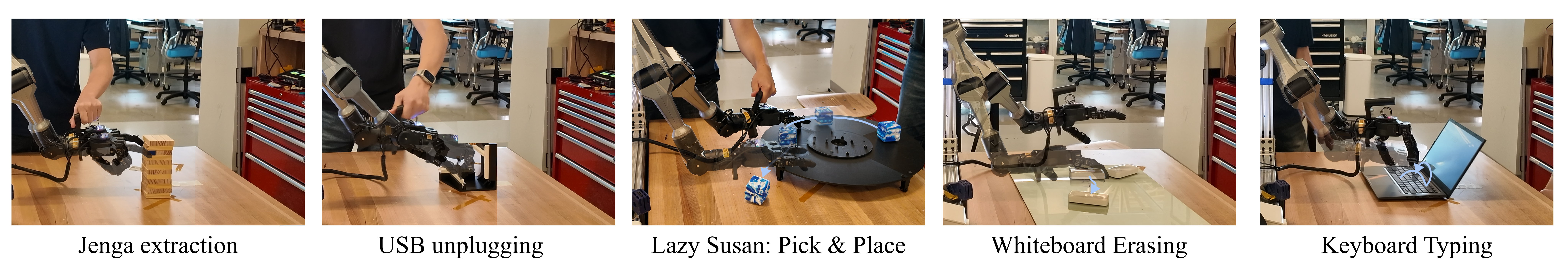}
\caption{\textbf{Evaluation tasks.}
\ACRO is evaluated on five contact-rich dexterous manipulation tasks: Jenga manipulation, USB unplugging, Lazy Susan pick-and-place, whiteboard erasing, and keyboard typing.}
 \label{fig:tasks}
\end{figure*}

\begin{figure}[t!]
 \centering
 \includegraphics[width=\linewidth]{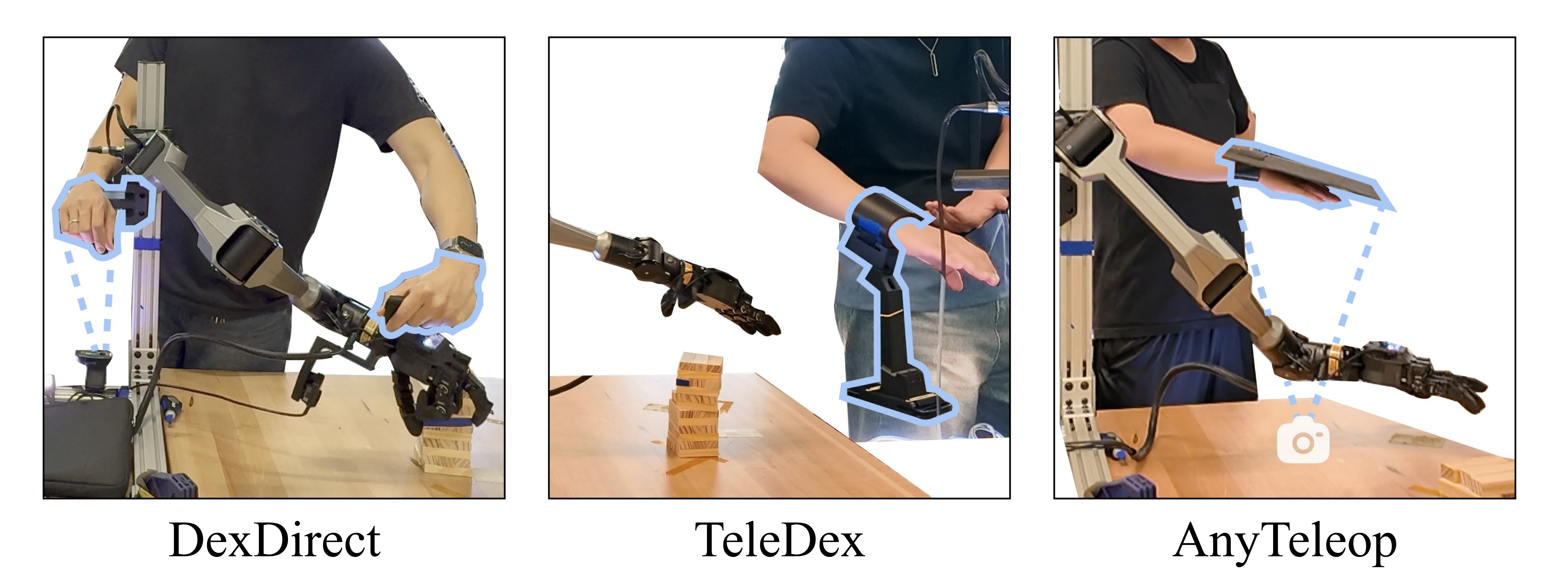}
\caption{\textbf{Comparison of operator interfaces for dexterous teleoperation.}
\ACRO combines direct physical guidance of the robot arm with vision-based hand retargeting, whereas TeleDex relies on a handheld control interface and AnyTeleop tracks the operator's arm and hand motion entirely through vision.}
 \label{fig:main}
\end{figure}

\subsection{Direct Kinesthetic Arm Guidance}
\label{sec:system:arm}

To compensate for the weight of the arm and end-effector payload while
retaining low resistance to manual motion, the arm operates in a
gravity-compensated drag mode with zero position stiffness and low
joint damping. During demonstration collection, the joint-level controller runs at
\SI{200}{\hertz}.
Before operation, we perform a staged system-identification procedure to
estimate accurate gravity direction, end-effector payload mass and center
of mass, joint-wise torque-delivery scaling, and Coulomb friction.
These robot-specific parameters are used to correct the gravity torques computed
from the MuJoCo model before applying them as feedforward torque commands.
With zero position gain, the controller does not pull the arm toward a
reference configuration, allowing the operator to directly position and
orient the physical robot arm.

Since the physical robot itself is guided, arm trajectories are recorded
directly from the joint encoders, and end-effector poses are obtained through
forward kinematics. No arm-side pose estimation, retargeting,
inverse-kinematics solver, or Cartesian motion generator is required during
collection. Each recorded configuration therefore corresponds to a state
physically attained by the robot, subject to its joint and controller limits.

The rigid mechanical connection also transmits interaction
force through the handle, providing the operator with passive contact
cues without a separate haptic device. Joint motion remains bounded by the
robot's hardware and controller safety limits, and an emergency stop is
available throughout operation.

\subsection{Vision-Based Finger Retargeting}
\label{sec:system:hand}

A palm-facing webcam captures the operator's right hand at
\SI{30}{\hertz}. Each RGB frame is processed by
MediaPipe Hands~\cite{mediapipe}, which detects 21 three-dimensional
hand landmarks.

The wrist-relative landmarks are mapped to the MIDAS hand using the
optimization-based \texttt{dex-retargeting} framework released with
AnyTeleop~\cite{anyteleop}. At each time step, the optimizer computes
robot joint targets by matching selected human and robot finger
vectors subject to the robot joint limits.

No global wrist position or orientation estimated from the webcam is
used for robot control, because the robot arm independently
determines the global pose of the hand. This avoids relying on
monocular metric wrist-pose estimation and confines the vision
pipeline to hand-local finger articulation.

The forearm support helps keep the palm centered in the camera view
and limits large wrist motion while leaving the fingers free to
articulate. The tracked hand therefore does not need to reproduce the global
motion or orientation of the robot hand.

\subsection{Multimodal Demonstration Recording}
\label{sec:system:recording}

Arm joint states, hand actuator states, tactile measurements, and
wrist-camera images are timestamped using a common monotonic clock on
the logging computer. Arm and hand states are logged at
\SI{200}{\hertz}, tactile measurements at \SI{60}{\hertz}, and
wrist-camera images at \SI{30}{\hertz}.

For policy training, the proprioceptive and tactile streams are
aligned to the wrist-camera timestamps to form time-aligned
\SI{30}{\hertz} demonstration episodes. For each camera timestamp,
we select the nearest MIDAS hand encoder, PIPER end-effector pose,
and tactile samples using nearest-neighbor timestamp matching,
without temporal interpolation.

\subsection{Policy Architecture}
\label{sec:system:recording}

Our wrist-camera RGB frame is downscaled to $240\times240$, with a random $224\times224$ crop and color jitter during training and a center crop during inference. A DINOv2 ViT-S/14 encoder~\cite{dinov2} turns this frame into an embedding that acts as the first part of the condition. In addition to the visual embedding, we append the 13-D vector of the absolute hand positions, along with a 4-D reading from the PaXini tactile sensors on the MIDAS Hand~\cite{midas}: mean thumb, max thumb, mean index, and max index. The policy emits a 19-D action as its command: relative positions for the 6-DoF end-effector target and 13 finger targets.

\section{Experiments and Results}
\label{sec:Results}

\begin{table*}[t]
    \centering
    \caption{
        Summary of quantitative results in the user study
        (mean $\pm$ SEM).
        Bold indicates the best performance for each metric.
    }
    \label{tab:user_study_results}
    \resizebox{\textwidth}{!}{%
    \setlength{\tabcolsep}{3pt}
    \begin{tabular}{lcccccccccc}
        \toprule
        & \multicolumn{2}{c}{\textbf{Jenga extraction}}
        & \multicolumn{2}{c}{\textbf{USB unplugging}}
        & \multicolumn{2}{c}{\makecell{\textbf{Lazy Susan rotation}\\\textbf{and pick-and-place}}}
        & \multicolumn{2}{c}{\textbf{Whiteboard erasing}}
        & \multicolumn{2}{c}{\textbf{Keyboard typing}} \\

        \cmidrule(lr){2-3}
        \cmidrule(lr){4-5}
        \cmidrule(lr){6-7}
        \cmidrule(lr){8-9}
        \cmidrule(lr){10-11}

        \textbf{Method}
        & \textbf{Success Rate} & \textbf{Time (s)$^{\dagger}$}
        & \textbf{Success Rate} & \textbf{Time (s)$^{\dagger}$}
        & \textbf{Success Rate} & \textbf{Time (s)$^{\dagger}$}
        & \textbf{Success Rate} & \textbf{Time (s)$^{\dagger}$}
        & \textbf{Success Rate} & \textbf{Time (s)$^{\dagger}$} \\
        \midrule

        DexDirect
        & $\mathbf{0.94 \pm 0.02}$
        & $\mathbf{10.46 \pm 0.82}$
        & $\mathbf{0.95 \pm 0.02}$
        & $\mathbf{5.88 \pm 0.65}$
        & $\mathbf{0.96 \pm 0.02}$
        & $\mathbf{8.36 \pm 0.49}$
        & $\mathbf{0.91 \pm 0.04}$
        & $\mathbf{13.85 \pm 1.39}$
        & $\mathbf{0.71 \pm 0.04}$
        & $\mathbf{6.60 \pm 0.32}$ \\

        TeleDex
        & $0.69 \pm 0.08$
        & $49.76 \pm 5.98$
        & $0.81 \pm 0.04$
        & $10.36 \pm 1.90$
        & $0.85 \pm 0.06$
        & $21.96 \pm 2.73$
        & $0.56 \pm 0.12$
        & $39.10 \pm 7.82$
        & $0.32 \pm 0.06$
        & $17.50 \pm 3.97$ \\

        AnyTeleop
        & $0.44 \pm 0.12$
        & $35.50 \pm 5.42$
        & $0.39 \pm 0.09$
        & $17.75 \pm 5.74$
        & $0.50 \pm 0.13$
        & $47.14 \pm 11.15$
        & $0.00 \pm 0.00$
        & ---
        & $0.00 \pm 0.00$
        & --- \\

        \bottomrule
    \end{tabular}%
    }

    \vspace{2pt}
    \begin{minipage}{0.99\textwidth}
        \footnotesize
        $^{\dagger}$ Completion time was computed over successful task
        executions. A dash indicates that no successful executions were
        available for computing completion time.
    \end{minipage}
\end{table*}


We evaluate DexDirect against two published low-setup
teleoperation systems, AnyTeleop and
TeleDex. Our evaluation addresses two primary
questions: whether DexDirect increases the number of successful
demonstrations collected under a fixed time budget, and whether it
reduces the workload imposed on the operator. We additionally verify
that demonstrations collected with DexDirect can be used for
downstream policy learning.

\subsection{Baseline Implementations}
\label{sec:exp:baselines}
\noindent\textbf{AnyTeleop.}
We reproduced the arm-control pathway of AnyTeleop
on the same PIPER--MIDAS platform. An Intel RealSense RGB-D camera
was used to estimate the operator's wrist pose following the
pose-estimation procedure described by AnyTeleop. The estimated wrist
pose was transformed into a Cartesian end-effector target in the robot
base frame. Hand-pose detection and end-effector target updates were
performed at \SI{30}{\hertz}. Following AnyTeleop's CuRobo-based
motion-generation pipeline, the target poses were passed to CuRobo's
receding-horizon \texttt{MpcSolver}, which ran at
\SI{120}{\hertz} on a separate workstation equipped with an NVIDIA
RTX~4070~Ti. The resulting joint-space trajectories were streamed to
the robot's low-level controller. To improve hand tracking reliability, the operator wore a black lightweight plate on the back of the hand.

\noindent\textbf{TeleDex.}
We implemented the TeleDex baseline using the authors' publicly released open-source software and hardware.

\subsection{User Study Design}
\label{sec:experiments:user-study}

\noindent\textbf{Participants and interfaces.}
We conducted a user study with ten participants who had no
prior experience with DexDirect and both basline methods. Each participant operated the same
PIPER--MIDAS robot using three interfaces: DexDirect, TeleDex, and
AnyTeleop. Across conditions, the robot arm, dexterous hand,
finger-retargeting pipeline, visual feedback, and task environments
were held fixed; the interfaces differed in how the operator commanded
the global motion of the robot hand.

The interface order was counterbalanced across participants using the
cyclic orders ABC, BCA, and CAB. Each participant completed the full
five-task sequence with one interface before proceeding to the next
interface.

Before the timed evaluation of each task, participants were
given one minute to practice the task with the corresponding interface. 
For each task, participants were instructed to complete as many
successful repetitions as possible within a fixed time budget.
The budget was \SI{120}{\second} for each task except Jenga, for which
participants received \SI{180}{\second}. Each completed attempt was
classified as successful or unsuccessful according to the
task-specific criteria described below.

\noindent\textbf{Tasks.}
The five tasks shown in Fig.~\ref{fig:tasks} were selected to
span distinct arm-control requirements, including large-workspace
positioning, sustained contact-force regulation, discrete precision
contact, and constrained extraction.

\begin{itemize}
    \item \textit{Lazy Susan rotation and pick-and-place} evaluates
    sustained tangential contact, workspace reconfiguration, and global
    arm positioning. The target cube is initially placed on the side of
    the platform farthest from the operator and robot hand. The participant
    first rotates the Lazy Susan by approximately $180^\circ$ through
    direct contact to bring the cube into the near workspace, then grasps
    the cube and places it in the designated target region. A repetition
    is successful when the platform is rotated as prescribed and the cube
    is transferred to the target region without being dropped.

    \item \textit{Whiteboard erasing} requires simultaneous tangential
    motion along the board and regulation of normal contact force.
    The participant erases the designated marking while maintaining
    contact between the eraser and the board. A repetition is successful
    when the word `robot' in the prescribed region is fully erased.

    \item \textit{Keyboard typing} evaluates repeated discrete contact
    with small spatial targets. The participant presses the prescribed
    key sequence `bot'. A repetition is successful when all requested
    keys are registered in the correct order without an incorrect key
    press.

    \item \textit{Jenga extraction} combines precise hand placement,
    stable fingertip contact, and gradual regulation of extraction
    force. The participant pinches and completely removes the designated
    block. A repetition is successful when the target block is extracted
    without collapsing the stack.

    \item \textit{USB unplugging} requires a stable grasp, orientation
    alignment with the connector axis, and directed extraction force.
    A repetition is successful when the connector is fully removed from
    the socket.
\end{itemize}

An attempt was marked as unsuccessful if the success criterion was not
met, an irreversible task error occurred, or the allotted time elapsed
before completion.

\subsection{Demonstration Throughput and Task Performance}
\label{sec:experiments:throughput}

Under the same time budget per-condition, \ACRO collected
481 successful demonstrations across participants and tasks, compared
with 151 for TeleDex and 28 for AnyTeleop. This corresponds to
$3.2\times$ the successful-demonstration throughput of TeleDex and
$17.2\times$ that of AnyTeleop.

As summarized in Table~\ref{tab:user_study_results}, \ACRO achieved the
highest success rate on every task. Its success rate ranged from
0.71 on keyboard typing to 0.96 on Lazy Susan rotation and pick-and-place,
whereas TeleDex ranged from 0.32 to 0.85 and AnyTeleop from
0 to 0.50. In particular, AnyTeleop produced no successful whiteboard or
keyboard demonstrations within the evaluation budget. Video review
showed that participants struggled to regulate the end-effector depth
normal to the whiteboard, resulting in insufficient or unstable
contact during erasing. During keyboard typing, the delayed robot response made brief press-and-release motions difficult to
execute, frequently causing repeated registrations of the same key.

\ACRO also reduced the time required for successful task
executions. Relative to TeleDex, it achieved completion-time speedups
of $4.8\times$ on Jenga, $1.8\times$ on USB unplugging,
$2.6\times$ on Lazy Susan, $2.8\times$ on whiteboard erasing,
and $2.7\times$ on keyboard typing. Relative to AnyTeleop,
the speedups on tasks for which AnyTeleop produced successful trials
were $3.4\times$ on Jenga, $3.0\times$ on USB unplugging, and
$5.6\times$ on Lazy Susan.

These results indicate that DexDirect's throughput advantage is not
caused solely by allowing more unsuccessful attempts. Participants
both completed individual trials more quickly and succeeded on a
larger fraction of those attempts.

\begin{figure}[t!]
 \centering
 \includegraphics[width=\linewidth]{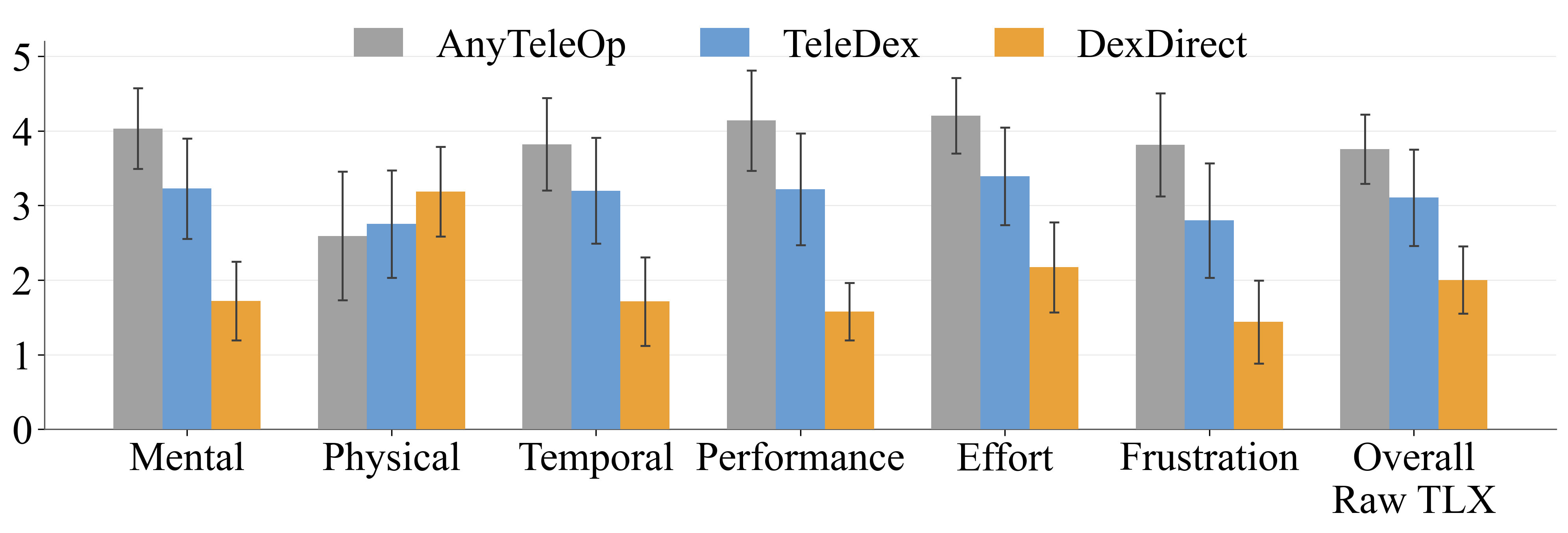}
\caption{Mean adapted NASA--TLX scores for AnyTeleop, TeleDex, and DexDirect, averaged across ten participants and five tasks. Error bars indicate 95\% confidence intervals computed from participant-level scores. Lower scores indicate lower perceived workload. The performance item is reverse-coded so that higher values consistently indicate poorer perceived performance.}
 \label{fig:result}
\end{figure}

\subsection{Operator Workload}
\label{sec:experiments:workload}

\ACRO substantially reduced subjective workload despite requiring
the operator to physically move the robot arm. The adapted Raw TLX score
is the unweighted mean of six workload dimensions, with lower scores
indicating lower perceived workload. The performance item was
reverse-coded so that higher values consistently indicate poorer
perceived performance and greater workload. \ACRO achieved an
overall Raw TLX score of 2.01, compared with 3.08 for TeleDex and 3.81
for AnyTeleop. Paired Wilcoxon signed-rank tests on participant-level scores found a
significant reduction relative to both AnyTeleop ($p=0.00195$) and
TeleDex ($p=0.01953$). Both values are below the conventional
significance threshold of $0.05$, indicating that the observed
reductions are unlikely to arise from random participant-level
variation alone.

The largest differences occurred in cognitive dimensions. Relative to
TeleDex, \ACRO reduced mental demand from 3.19 to 1.77,
frustration from 2.79 to 1.47, and effort from 3.40 to 2.17.
Temporal demand and perceived-performance workload were also lower.
Physical demand was the only dimension for which \ACRO performed worse, increasing from 2.72 for TeleDex and 2.62 for AnyTeleop to
3.23.

The workload profile therefore exposes the central trade-off of direct
kinesthetic guidance: it replaces cognitive effort associated with
tracking, coordinate remapping, and visual correction with the
physical effort of moving the gravity-compensated robot. The lower
overall workload suggests that this trade is favorable for repeated
demonstration collection.

\subsection{Policy Training}
To evaluate whether DexDirect is suitable for end-to-end policy learning, we design a simple cube pick-and-place task. The robot picks up a cube from the table and drops it onto a box located \SI{30}{\centi\meter} away. Success is defined as lifting the cube, transporting it above the box, and releasing it such that it lands on the box.





For policy learning, a single operator separately collects
200 demonstrations using DexDirect. These demonstrations are
collected independently of the user study. We train a diffusion policy for the cube pick-and-place task and evaluate the policy over 20 episodes without operator intervention, using the success definitions described above.

The cube pick-and-place task achieves a success rate of 90\% (18/20) over 20 evaluation episodes.




\begin{figure}[t!]
 \centering
 \includegraphics[width=\linewidth]{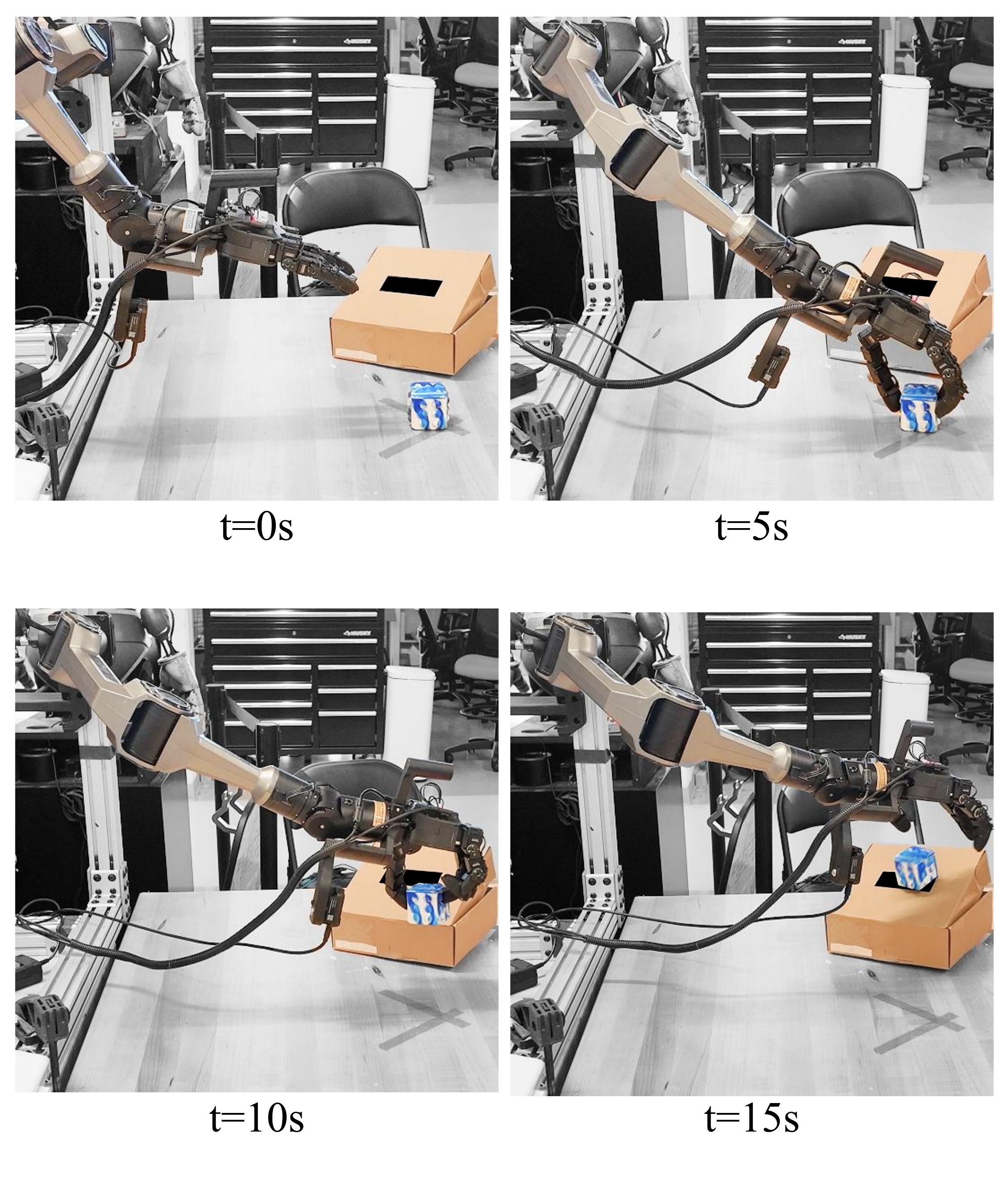}
    \caption{\textbf{Representative policy rollouts.}
    The learned policy grasps a cube from a randomized initial pose, lifts it, and places it onto the target box.
    Each sequence shows the rollout at $t=\{0,5,10,15\}$\,s.}
 \label{fig:result}
\end{figure}

\section{Limitations}

\ACRO improves demonstration throughput through direct physical arm
guidance, but introduces several trade-offs.

First, the system is inherently co-located and requires a robot capable of
safe, low-impedance, gravity-compensated guidance. It therefore does not
support remote teleoperation and may be less ergonomic for larger or
higher-inertia manipulators. Consistent with this trade-off, DexDirect
increased physical demand despite reducing overall workload.

Second, the current interface uses both operator hands to control one
arm-hand system, preventing direct single-operator bimanual control. A
possible extension is to mechanically couple each operator hand to one
robot arm while tracking the fingers of the same hand, although its
visibility, ergonomics, and safety remain to be studied.

Third, finger control relies on monocular landmark detection and
morphology-dependent retargeting, and is therefore still susceptible to
tracking and retargeting errors. The interface also does not provide
distributed fingertip force feedback.

Finally, our evaluation is limited to ten participants, five short tasks,
one robot platform, and a policy-learning task. It also evaluates the
combined effect of direct coupling without isolating coordinate remapping,
latency, and passive contact cues. Future work will consider longer-term
studies, additional platforms, bimanual operation, and controlled
ablations of these factors.
\section{Conclusion}

We presented DexDirect, a low-setup interface that combines direct kinesthetic arm guidance with vision-based finger retargeting for efficient dexterous demonstration collection. In a ten-participant study across five tasks, DexDirect collected 481 successful demonstrations, compared with 151 using TeleDex and 28 using AnyTeleop, while reducing completion time and overall workload. Policy trained on DexDirect demonstrations achieved 90\% success on cube pick-and-place. These results show that eliminating the arm-side tracking and retargeting loop provides a practical and effective path toward scalable dexterous robot learning.



\renewcommand*{\bibfont}{\footnotesize}
\printbibliography

@inproceedings{anyteleop,
  title     = {AnyTeleop: A General Vision-Based Dexterous Robot Arm-Hand Teleoperation System},
  author    = {Qin, Yuzhe and Yang, Wei and Huang, Binghao and Van Wyk, Karl and Su, Hao and Wang, Xiaolong and Chao, Yu-Wei and Fox, Dieter},
  booktitle = {Robotics: Science and Systems},
  year      = {2023}
}

@inproceedings{dexpilot,
  title        = {DexPilot: Vision-Based Teleoperation of Dexterous Robotic Hand-Arm System},
  author       = {Handa, Ankur and Van Wyk, Karl and Yang, Wei and Liang, Jacky and Chao, Yu-Wei and Wan, Qian and Birchfield, Stan and Ratliff, Nathan and Fox, Dieter},
  booktitle    = {IEEE International Conference on Robotics and Automation (ICRA)},
  year         = {2020}
}

@misc{teledex,
      title={TeleDex: Accessible Dexterous Teleoperation}, 
      author={Omar Rayyan and Maximilian Gilles and Yuchen Cui},
      year={2026},
      eprint={2603.17065},
      archivePrefix={arXiv},
      primaryClass={cs.RO},
      url={https://arxiv.org/abs/2603.17065}, 
}

@misc{dexexo,
      title={DexEXO: A Wearability-First Dexterous Exoskeleton for Operator-Agnostic Demonstration and Learning}, 
      author={Alvin Zhu and Mingzhang Zhu and Beom Jun Kim and Jose Victor S. H. Ramos and Yike Shi and Yufeng Wu and Raayan Dhar and Fuyi Yang and Ruochen Hou and Hanzhang Fang and Quanyou Wang and Yuchen Cui and Dennis W. Hong},
      year={2026},
      eprint={2603.17323},
      archivePrefix={arXiv},
      primaryClass={cs.RO},
      url={https://arxiv.org/abs/2603.17323}, 
}

@inproceedings{dexumi,
  title={DexUMI: Using Human Hand as the Universal Manipulation Interface for Dexterous Manipulation},
  author={Xu, Mengda and Zhang, Han and Hou, Yifan and Xu, Zhenjia and Fan, Linxi and Veloso, Manuela and Song, Shuran},
  booktitle={Conference on Robot Learning},
  pages={437--459},
  year={2025},
  organization={PMLR}
}

@inproceedings{opentelevision,
  title        = {Open-TeleVision: Teleoperation with Immersive Active Visual Feedback},
  author       = {Cheng, Xuxin and Li, Jialong and Yang, Shiqi and Yang, Ge and Wang, Xiaolong},
  booktitle    = {Conference on Robot Learning (CoRL)},
  year         = {2024},
  eprint       = {2407.01512},
  archivePrefix= {arXiv},
  primaryClass = {cs.RO},
  url          = {https://arxiv.org/abs/2407.01512}
}

@misc{bunnyvisionpro,
  title        = {Bunny-VisionPro: Real-Time Bimanual Dexterous Teleoperation for Imitation Learning},
  author       = {Ding, Runyu and Qin, Yuzhe and Zhu, Jiyue and Jia, Chengzhe and Yang, Shiqi and Yang, Ruihan and Qi, Xiaojuan and Wang, Xiaolong},
  year         = {2024},
  eprint       = {2407.03162},
  archivePrefix= {arXiv},
  primaryClass = {cs.RO},
  url          = {https://arxiv.org/abs/2407.03162}
}

@misc{telekinesis,
  title        = {Robotic Telekinesis: Learning a Robotic Hand Imitator by Watching Humans on YouTube},
  author       = {Sivakumar, Aravind and Shaw, Kenneth and Pathak, Deepak},
  booktitle    = {Robotics: Science and Systems (RSS)},
  year         = {2022},
  eprint       = {2202.10448},
  archivePrefix= {arXiv},
  primaryClass = {cs.RO},
  url          = {https://arxiv.org/abs/2202.10448}
}

@inproceedings{omnih2o,
  title        = {OmniH2O: Universal and Dexterous Human-to-Humanoid Whole-Body Teleoperation and Learning},
  author       = {He, Tairan and Luo, Zhengyi and He, Xialin and Xiao, Wenli and Zhang, Chong and Zhang, Weinan and Kitani, Kris and Liu, Changliu and Shi, Guanya},
  booktitle    = {Conference on Robot Learning (CoRL)},
  year         = {2024},
  eprint       = {2406.08858},
  archivePrefix= {arXiv},
  primaryClass = {cs.RO},
  url          = {https://arxiv.org/abs/2406.08858}
}

@article{wrede2013kinesthetic,
  author  = {Sebastian Wrede and Christian Emmerich and Ricarda Gr{\"u}nberg
             and Arne Nordmann and Agnes Swadzba and Jochen J. Steil},
  title   = {A User Study on Kinesthetic Teaching of Redundant Robots
             in Task and Configuration Space},
  journal = {Journal of Human-Robot Interaction},
  volume  = {2},
  number  = {1},
  pages   = {56--81},
  year    = {2013},
  doi     = {10.5898/JHRI.2.1.Wrede}
}

@inproceedings{akgun2012kinesthetic,
  author    = {Baris Akgun and Maya Cakmak and Jae Wook Yoo
               and Andrea L. Thomaz},
  title     = {Trajectories and Keyframes for Kinesthetic Teaching:
               A Human-Robot Interaction Perspective},
  booktitle = {Proceedings of the ACM/IEEE International Conference
               on Human-Robot Interaction (HRI)},
  pages     = {391--398},
  year      = {2012},
  publisher = {ACM},
  doi       = {10.1145/2157689.2157815}
}

@inproceedings{li2025demonstration,
  author    = {Haozhuo Li and Yuchen Cui and Dorsa Sadigh},
  title     = {How to Train Your Robots? The Impact of Demonstration
               Modality on Imitation Learning},
  booktitle = {Proceedings of the IEEE International Conference on
               Robotics and Automation (ICRA)},
  pages     = {1113--1120},
  year      = {2025},
  doi       = {10.1109/ICRA55743.2025.11128520}
}

@incollection{hart1988development,
  author    = {Hart, Sandra G. and Staveland, Lowell E.},
  title     = {Development of NASA-TLX (Task Load Index):
               Results of Empirical and Theoretical Research},
  booktitle = {Human Mental Workload},
  editor    = {Hancock, Peter A. and Meshkati, Najmedin},
  pages     = {139--183},
  publisher = {North-Holland},
  year      = {1988},
  doi       = {10.1016/S0166-4115(08)62386-9}
}

@misc{openteach,
  title        = {Open Teach: A Versatile Teleoperation System for Robotic Manipulation},
  author       = {Iyer, Aadhithya and Peng, Zhuoran and Dai, Yinlong and Guzey, Irmak and Haldar, Siddhant and Chintala, Soumith and Pinto, Lerrel},
  year         = {2024},
  eprint       = {2403.07870},
  archivePrefix= {arXiv},
  primaryClass = {cs.RO},
  url          = {https://arxiv.org/abs/2403.07870}
}

@inproceedings{gello,
  title        = {GELLO: A General, Low-Cost, and Intuitive Teleoperation Framework for Robot Manipulators},
  author       = {Wu, Philipp and Shentu, Yide and Yi, Zhongke and Lin, Xingyu and Abbeel, Pieter},
  booktitle    = {IEEE/RSJ International Conference on Intelligent Robots and Systems (IROS)},
  year         = {2024},
  eprint       = {2309.13037},
  archivePrefix= {arXiv},
  primaryClass = {cs.RO},
  url          = {https://arxiv.org/abs/2309.13037}
}

@inproceedings{aloha,
  title        = {Learning Fine-Grained Bimanual Manipulation with Low-Cost Hardware},
  author       = {Zhao, Tony Z. and Kumar, Vikash and Levine, Sergey and Finn, Chelsea},
  booktitle    = {Robotics: Science and Systems (RSS)},
  year         = {2023},
  eprint       = {2304.13705},
  archivePrefix= {arXiv},
  primaryClass = {cs.RO},
  url          = {https://arxiv.org/abs/2304.13705}
}

@inproceedings{ace,
  title        = {ACE: A Cross-Platform Visual-Exoskeletons System for Low-Cost Dexterous Teleoperation},
  author       = {Yang, Shiqi and Liu, Minghuan and Qin, Yuzhe and Ding, Runyu and Li, Jialong and Cheng, Xuxin and Yang, Ruihan and Yi, Sha and Wang, Xiaolong},
  booktitle    = {Conference on Robot Learning (CoRL)},
  year         = {2024},
  eprint       = {2408.11805},
  archivePrefix= {arXiv},
  primaryClass = {cs.RO},
  url          = {https://arxiv.org/abs/2408.11805}
}

@misc{dexop,
  title        = {DEXOP: A Device for Robotic Transfer of Dexterous Human Manipulation},
  author       = {Fang, Hao-Shu and Romero, Branden and Xie, Yichen and Hu, Arthur and Huang, Bo-Ruei and Alvarez, Juan and Kim, Matthew and Margolis, Gabriel and Anbarasu, Kavya and Tomizuka, Masayoshi and Adelson, Edward and Agrawal, Pulkit},
  year         = {2025},
  eprint       = {2509.04441},
  archivePrefix= {arXiv},
  primaryClass = {cs.RO},
  url          = {https://arxiv.org/abs/2509.04441}
}

@misc{exostart,
  title        = {ExoStart: Efficient Learning for Dexterous Manipulation with Sensorized Exoskeleton Demonstrations},
  author       = {Si, Zilin and Chen, Jose Enrique and Karagozler, M. Emre and Bronars, Antonia and Hutchinson, Jonathan and Lampe, Thomas and Gileadi, Nimrod and Howell, Taylor and Saliceti, Stefano and Barczyk, Lukasz and Correa, Ilan Olivarez and Erez, Tom and Shridhar, Mohit and Martins, Murilo Fernandes and Bousmalis, Konstantinos and Heess, Nicolas and Nori, Francesco and Bauza, Maria},
  year         = {2025},
  eprint       = {2506.11775},
  archivePrefix= {arXiv},
  primaryClass = {cs.RO},
  url          = {https://arxiv.org/abs/2506.11775}
}

@misc{globallocal,
  title        = {Global-Local Interface for On-Demand Teleoperation},
  author       = {Zhou, Jianshu and Liang, Boyuan and Huang, Junda and Zhang, Ian and Tomizuka, Masayoshi},
  year         = {2025},
  eprint       = {2502.09960},
  archivePrefix= {arXiv},
  primaryClass = {cs.RO},
  url          = {https://arxiv.org/abs/2502.09960}
}

@inproceedings{umi,
  title        = {Universal Manipulation Interface: In-The-Wild Robot Teaching Without In-The-Wild Robots},
  author       = {Chi, Cheng and Xu, Zhenjia and Pan, Chuer and Cousineau, Eric and Burchfiel, Benjamin and Feng, Siyuan and Tedrake, Russ and Song, Shuran},
  booktitle    = {Robotics: Science and Systems (RSS)},
  year         = {2024},
  eprint       = {2402.10329},
  archivePrefix= {arXiv},
  primaryClass = {cs.RO},
  url          = {https://arxiv.org/abs/2402.10329}
}

@inproceedings{dexcap,
  title        = {DexCap: Scalable and Portable Mocap Data Collection System for Dexterous Manipulation},
  author       = {Wang, Chen and Shi, Haochen and Wang, Weizhuo and Zhang, Ruohan and Fei-Fei, Li and Liu, C. Karen},
  booktitle    = {Robotics: Science and Systems (RSS)},
  year         = {2024},
  eprint       = {2403.07788},
  archivePrefix= {arXiv},
  primaryClass = {cs.RO},
  url          = {https://arxiv.org/abs/2403.07788}
}

@article{dexforce,
  title        = {DexForce: Extracting Force-informed Actions from Kinesthetic Demonstrations for Dexterous Manipulation},
  author       = {Chen, Claire and Yu, Zhongchun and Choi, Hojung and Cutkosky, Mark and Bohg, Jeannette},
  journal      = {IEEE Robotics and Automation Letters (RA-L)},
  year         = {2025},
  eprint       = {2501.10356},
  archivePrefix= {arXiv},
  primaryClass = {cs.RO},
  url          = {https://arxiv.org/abs/2501.10356}
}

@inproceedings{forcemimic,
  title        = {ForceMimic: Force-Centric Imitation Learning with Force-Motion Capture System for Contact-Rich Manipulation},
  author       = {Liu, Wenhai and Wang, Junbo and Wang, Yiming and Wang, Weiming and Lu, Cewu},
  booktitle    = {IEEE International Conference on Robotics and Automation (ICRA)},
  year         = {2025},
  eprint       = {2410.07554},
  archivePrefix= {arXiv},
  primaryClass = {cs.RO},
  url          = {https://arxiv.org/abs/2410.07554}
}

@misc{dexmouse,
  title        = {DEX-Mouse: A Low-cost Portable and Universal Interface with Force Feedback for Data Collection of Dexterous Robotic Hands},
  author       = {Koh, Joonho and Jung, Haechan and Kim, Nayoung and Ko, Wook and Nam, Changjoo},
  year         = {2026},
  eprint       = {2604.15013},
  archivePrefix= {arXiv},
  primaryClass = {cs.RO},
  url          = {https://arxiv.org/abs/2604.15013}
}

@misc{glovity,
  title        = {Glovity: Learning Dexterous Contact-Rich Manipulation via Spatial Wrench Feedback Teleoperation System},
  author       = {Gao, Yuyang and Ma, Haofei and Zheng, Pai},
  year         = {2025},
  eprint       = {2510.09229},
  archivePrefix= {arXiv},
  primaryClass = {cs.RO},
  url          = {https://arxiv.org/abs/2510.09229}
}

@misc{midas,
  title        = {MIDAS Hand: Modular low-Impedance Direct-drive Anthropomorphic Sensing Hand},
  author       = {Zhu, Alvin and Zhu, Mingzhang and Kim, Beom Jun and Wang, Quanyou and Ramos, Jose Victor S. H. and Hong, Dennis},
  year         = {2026},
  eprint       = {2607.14487},
  archivePrefix= {arXiv},
  primaryClass = {cs.RO},
  url          = {https://arxiv.org/abs/2607.14487}
}

@article{dinov2,
  title        = {DINOv2: Learning Robust Visual Features without Supervision},
  author       = {Oquab, Maxime and Darcet, Timoth\'ee and Moutakanni, Th\'eo and Vo, Huy and Szafraniec, Marc and Khalidov, Vasil and Fernandez, Pierre and Haziza, Daniel and Massa, Francisco and El-Nouby, Alaaeldin and Assran, Mahmoud and Ballas, Nicolas and Galuba, Wojciech and Howes, Russell and Huang, Po-Yao and Li, Shang-Wen and Misra, Ishan and Rabbat, Michael and Sharma, Vasu and Synnaeve, Gabriel and Xu, Hu and J\'egou, Herv\'e and Mairal, Julien and Labatut, Patrick and Joulin, Armand and Bojanowski, Piotr},
  journal      = {Transactions on Machine Learning Research (TMLR)},
  year         = {2024},
  eprint       = {2304.07193},
  archivePrefix= {arXiv},
  primaryClass = {cs.CV},
  url          = {https://arxiv.org/abs/2304.07193}
}

@misc{mediapipe,
  title        = {MediaPipe Hands: On-device Real-time Hand Tracking},
  author       = {Zhang, Fan and Bazarevsky, Valentin and Vakunov, Andrey and Tkachenka, Andrei and Sung, George and Chang, Chuo-Ling and Grundmann, Matthias},
  year         = {2020},
  eprint       = {2006.10214},
  archivePrefix= {arXiv},
  primaryClass = {cs.CV},
  url          = {https://arxiv.org/abs/2006.10214}
}

@misc{lerobot,
  title        = {LeRobot: State-of-the-art Machine Learning for Real-World Robotics in Pytorch},
  author       = {Cadene, Remi and Alibert, Simon and Soare, Alexander and Gallouedec, Quentin and Zouitine, Adil and Palma, Steven and Kooijmans, Pepijn and Aractingi, Michel and Shukor, Mustafa and Aubakirova, Dana and Russi, Martino and Capuano, Francesco and Pascal, Caroline and Choghari, Jade and Meftah, Khalil and Ellerbach, Maxime and Moss, Jess and Wolf, Thomas},
  howpublished = {\url{https://github.com/huggingface/lerobot}},
  year         = {2024}
}

@inproceedings{mobilealoha,
  title        = {Mobile ALOHA: Learning Bimanual Mobile Manipulation with Low-Cost Whole-Body Teleoperation},
  author       = {Fu, Zipeng and Zhao, Tony Z. and Finn, Chelsea},
  booktitle    = {Conference on Robot Learning (CoRL)},
  year         = {2024},
  eprint       = {2401.02117},
  archivePrefix= {arXiv},
  primaryClass = {cs.RO},
  url          = {https://arxiv.org/abs/2401.02117}
}

@article{lfd_survey,
  title        = {A survey of robot learning from demonstration},
  author       = {Argall, Brenna D. and Chernova, Sonia and Veloso, Manuela and Browning, Brett},
  journal      = {Robotics and Autonomous Systems},
  volume       = {57},
  number       = {5},
  pages        = {469--483},
  year         = {2009},
  doi          = {10.1016/j.robot.2008.10.024}
}

@article{forcematch,
  title        = {Multimodal and Force-Matched Imitation Learning with a See-Through Visuotactile Sensor},
  author       = {Ablett, Trevor and Limoyo, Oliver and Sigal, Adam and Jilani, Affan and Kelly, Jonathan and Siddiqi, Kaleem and Hogan, Francois and Dudek, Gregory},
  journal      = {IEEE Transactions on Robotics (T-RO)},
  volume       = {41},
  pages        = {946--959},
  year         = {2025},
  eprint       = {2311.01248},
  archivePrefix= {arXiv},
  primaryClass = {cs.RO},
  url          = {https://arxiv.org/abs/2311.01248}
}

@misc{kinedex,
  title        = {KineDex: Learning Tactile-Informed Visuomotor Policies via Kinesthetic Teaching for Dexterous Manipulation},
  author       = {Zhang, Di and Yuan, Chengbo and Wen, Chuan and Zhang, Hai and Zhao, Junqiao and Gao, Yang},
  year         = {2025},
  eprint       = {2505.01974},
  archivePrefix= {arXiv},
  primaryClass = {cs.RO},
  url          = {https://arxiv.org/abs/2505.01974}
}

@inproceedings{kinesoft,
  title        = {KineSoft: Learning Proprioceptive Manipulation Policies with Soft Robot Hands},
  author       = {Yoo, Uksang and Francis, Jonathan and Oh, Jean and Ichnowski, Jeffrey},
  booktitle    = {Conference on Robot Learning (CoRL)},
  year         = {2025},
  eprint       = {2503.01078},
  archivePrefix= {arXiv},
  primaryClass = {cs.RO},
  url          = {https://arxiv.org/abs/2503.01078}
}

@inproceedings{airexo2,
  title        = {AirExo-2: Scaling up Generalizable Robotic Imitation Learning with Low-Cost Exoskeletons},
  author       = {Fang, Hongjie and Wang, Chenxi and Wang, Yiming and Chen, Jingjing and Xia, Shangning and Lv, Jun and He, Zihao and Yi, Xiyan and Guo, Yunhan and Zhan, Xinyu and Yang, Lixin and Wang, Weiming and Lu, Cewu and Fang, Hao-Shu},
  booktitle    = {Conference on Robot Learning (CoRL)},
  year         = {2025},
  eprint       = {2503.03081},
  archivePrefix= {arXiv},
  primaryClass = {cs.RO},
  url          = {https://arxiv.org/abs/2503.03081}
}

@misc{nuexo,
  title        = {NuExo: A Wearable Exoskeleton Covering all Upper Limb ROM for Outdoor Data Collection and Teleoperation of Humanoid Robots},
  author       = {Zhong, Rui and Cheng, Chuang and Xu, Junpeng and Wei, Yantong and Guo, Ce and Zhang, Daoxun and Dai, Wei and Lu, Huimin},
  year         = {2025},
  eprint       = {2503.10554},
  archivePrefix= {arXiv},
  primaryClass = {cs.RO},
  url          = {https://arxiv.org/abs/2503.10554}
}

@misc{ume,
  title        = {Universal Manipulation Exoskeleton: Learning Compliant Whole-body Policies with Real-time Torque Feedback},
  author       = {Liang, Litian and Xu, Jingxi and Qi, Xinda and Cai, Yujun and Ding, Houzhu and Wang, Luqi and Sun, Zhixin and Chow, Jyh-Herng and Yang, Ming and Cutkosky, Mark},
  year         = {2026},
  eprint       = {2606.14218},
  archivePrefix= {arXiv},
  primaryClass = {cs.RO},
  url          = {https://arxiv.org/abs/2606.14218}
}

\end{document}